\documentclass{article}

\usepackage[utf8]{inputenc}
\usepackage[T1]{fontenc}

\PassOptionsToPackage{table, dvipsnames}{xcolor}
\usepackage{xcolor}

\PassOptionsToPackage{colorlinks=true, citecolor=blue, linkcolor=blue, urlcolor=black}{hyperref}

\usepackage{arxiv}
\usepackage[utf8]{inputenc}
\usepackage[T1]{fontenc}
\usepackage{url}
\usepackage{booktabs}
\usepackage{nicefrac}
\usepackage{microtype}
\usepackage{lipsum}
\usepackage{graphicx}
\usepackage{natbib}
\usepackage{doi}
\usepackage{algorithm}
\usepackage{multirow}
\usepackage{multicol}
\usepackage{amsmath,amssymb,amsfonts}
\usepackage{amsthm}
\usepackage{mathrsfs}
\usepackage{float}
\usepackage{algorithmic}
\usepackage{bm}
\usepackage{caption}
\usepackage{hyperref}

\title{Diffusion priors enhanced velocity model building from time-lag images using a neural operator}

\author{ \href{https://orcid.org/0009-0009-0709-9158}{\includegraphics[scale=0.06]{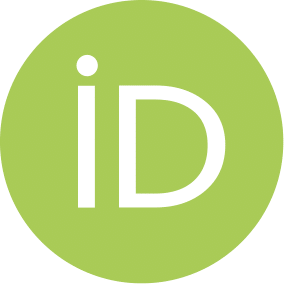}\hspace{1mm}Xiao~Ma}\\
	Division of Physical Science and Engineering\\
	King Abdullah University of Science and Technology\\
	Thuwal 23955-6900, Saudi Arabia \\
	\texttt{xiao.ma@kaust.edu.sa} \\
        \And
	\href{https://orcid.org/0000-0002-6336-7614}{\includegraphics[scale=0.06]{orcid.png}\hspace{1mm}Mohammad Hasyim Taufik} \\
	Division of Physical Science and Engineering\\
	King Abdullah University of Science and Technology\\
	Thuwal 23955-6900, Saudi Arabia \\
	\texttt{mohammad.taufik@kaust.edu.sa} \\
            \And
	\href{https://orcid.org/0000-0002-9363-9799}{\includegraphics[scale=0.06]{orcid.png}\hspace{1mm}Tariq~Alkhalifah} \\
	Division of Physical Science and Engineering\\
	King Abdullah University of Science and Technology\\
	Thuwal 23955-6900, Saudi Arabia \\
	\texttt{tariq.alkhalifah@kaust.edu.sa} \\
}

\renewcommand{\shorttitle}{Diffusion prior enhanced VMB using Neural Operator}
\graphicspath{{./Figures/}}
\hypersetup{
pdftitle={A template for the arxiv style},
pdfsubject={q-bio.NC, q-bio.QM},
pdfauthor={David S.~Hippocampus, Elias D.~Striatum},
pdfkeywords={First keyword, Second keyword, More},
}

\begin{document}
\maketitle

\begin{abstract}
Velocity model building serves as a crucial component for achieving high precision subsurface imaging. However, conventional velocity model building methods are often computationally expensive and time consuming. In recent years, with the rapid advancement of deep learning, particularly the success of generative models and neural operators, deep learning based approaches that integrate data and their statistics have attracted increasing attention in addressing the limitations of traditional methods. In this study, we propose a novel framework that combines generative models with neural operators to obtain high resolution velocity models efficiently. Within this workflow, the neural operator functions as a forward mapping operator to rapidly generate time lag reverse time migration (RTM) extended images from the true and migration velocity models. In this framework, the neural operator is acting as a surrogate for modeling followed by migration, which uses the true and migration velocities, respectively. The trained neural operator is then employed, through automatic differentiation, to gradually update the migration velocity placed in the true velocity input channel with high resolution components so that the output of the network matches the time lag images of observed data obtained using the migration velocity. By embedding a generative model, trained on a high-resolution velocity model distribution, which corresponds to the true velocity model distribution used to train the neural operator, as a regularizer, the resulting predictions are cleaner with higher resolution information. Both synthetic and field data experiments demonstrate the effectiveness of the proposed generative neural operator based velocity model building approach.
\end{abstract}

\keywords{Deep learning \and Neural operators \and Reverse time migration \and Diffusion model}
\section{\textbf{Introduction}}
Velocity model building (VMB) is a crucial component in geophysical exploration. An accurate subsurface velocity model enables precise seismic imaging, which serves as a foundation for subsequent applications such as oil and gas exploration \citep{xu2012full,adamczyk2014high}, carbon dioxide sequestration \citep{brossier2015velocity,lumley2019role}, and geothermal resource assessment \citep{hermans2014geophysical,kana2015review}. Conventional velocity model building methods are generally classified into ray-based and wave-equation-based approaches. The ray-tracing methods are based on the high-frequency asymptotic approximation of wave propagation, in which the seismic wavefield is approximated as energy traveling along ray paths. Within this theoretical framework, the propagation of wavefronts is governed by the Eikonal equation \citep{vidale1988finite,alkhalifah2002traveltime,waheed2013two,bin2015efficient,bin2021pinneik}, allowing the traveltime information to be inverted for subsurface velocity structures. Because of its computational efficiency and relative simplicity of implementation, ray-based velocity model building has been widely adopted in the geophysical industry \citep{barnes2011diving,lambare2014recent}. Nevertheless, ray-tracing methods have inherent limitations. The manual picking of velocity spectra introduces uncertainty, and the high-frequency assumption neglects finite-wavelength effects such as diffraction and scattering, making it difficult to accurately characterize complex wave phenomena. Consequently, ray-based approaches generally produce smooth velocity models that lack high-wavenumber details. \textcolor{black}{Alternatively, wave equation based methods have emerged,} directly utilizing both the amplitude and phase information of seismic wavefields to reconstruct the subsurface velocity model. Among these, migration velocity analysis (MVA) has demonstrated outstanding performance and has become one of the most powerful wave-equation-based techniques \citep{Alyahya1989Velocity,alkhalifah2003tau,sava2004wave,alali2020effectiveness}. MVA aims to iteratively update the subsurface velocity model such that seismic images are properly focused and reflector events are horizontally aligned after migration. Unlike travel-time tomography, which relies only on first-arrival information, MVA incorporates full reflection data to extract kinematic information from migrated gathers, making it capable of resolving complex velocity variations in depth. However, high-wavenumber details associated with reflectivity or small scale velocity perturbations are typically suppressed during the migration processes. As a result, the recovered velocity models usually lack sufficient resolution to capture the high wavenumber component. \textcolor{black}{Lately, full waveform inversion (FWI) has been used to obtain high resolution information of the subsurface \citep{tarantola1984inversion,choi2012application,warner2016adaptive}. In spite of its immense popularity, it is extremely costly to deploy, and it has its own set of challenges related to uniqueness and resolution \citep{alkhalifah2016full}}.\par

To provide new \textcolor{black}{avenues} for velocity model building, machine learning based approaches \citep{lecun2015deep,sandfort2019data,han2022survey} have recently gained significant attention. Specifically, inspired by the success of deep learning in computer vision \citep{long2015fully,he2016deep}, researchers have explored the use of Convolutional Neural Networks (CNNs) to directly infer high resolution velocity models from raw seismic data \citep{wu2019inversionnet,yang2019deep}. In such approaches, the network is trained on large datasets of synthetic seismic records to learn the complex nonlinear mapping from shot gathers to velocity models, enabling rapid velocity estimation during the inference stage. Despite their promising efficiency, purely data-driven methods suffer from several limitations. They typically lack physical interpretability, as the network does not explicitly incorporate the underlying wave propagation physics, and they exhibit limited generalization when applied to field data that differ from the training distribution \citep{kazei2021mapping}. To address these issues, recent studies have proposed integrating physical constraints or seismic imaging attributes into the learning process—for example, including seismic images \citep{lu2025seismic}, common image gathers \citep{geng2022deep,li2024deep}, or physics-guided loss functions \citep{ren2023physics}. These hybrid strategies enable the network to capture both high-wavenumber patterns and underlying physical relationships, thereby improving robustness and extending the applicability of machine learning based velocity model building to real seismic datasets \citep{harsuko2025synthesizing}. Although CNNs based methods have achieved promising results in velocity model building, the diverse and complex nature of the earth structure poses significant challenges to deep learning–based approaches. Therefore, more powerful deep learning algorithms are required to handle increasingly challenging scenarios. \par

Neural operators (NOs) and generative models, as emerging deep learning paradigms, have demonstrated remarkable performance across a wide range of applications \citep{kovachki2023neural,goswami2023physics,raonic2023convolutional,ma2024deepcache,bar2024lumiere}. In the geophysical domain, neural operators have already demonstrated promising performance in solving forward modeling problems. \cite{yang2021seismic} proposed a neural operator composed of Fourier Neural Operator (FNO) \citep{li2020fourier} layers to replace the conventional 2D acoustic wave equation operator, thereby enabling fast computation of time domain wavefields. By comparing the results with those obtained from traditional numerical simulations, they demonstrated that the neural operator \textcolor{black}{can be a surrogate for modeling.} Furthermore, they utilized reverse-mode automatic differentiation to compute the gradient of the neural operator predicted wavefield with respect to the velocity model, allowing the use of an FWI framework for velocity estimation. In addition to employing neural operators to learn time domain wave equations, several studies have also demonstrated their strong capability in modeling frequency domain wavefields. \cite{zou2025ambient} proposed using the UNO \citep{rahman2022u} to learn frequency domain surface wavefields. In their framework, the network takes the P-wave and S-wave velocity models as input, along with the source location, and outputs the corresponding frequency domain surface wavefield. After extensive data driven training, the UNO can accurately reproduce forward modeling results for various source positions. In the subsequent inversion stage based on neural operators, \textcolor{black}{ the inversion result obtained from the real surface wave data demonstrates that the neural operator based inversion produces result that are generally consistent with that obtained by conventional method.}  To inject more physical information into the inversion process,  \cite{huang2025physics} proposed incorporating a physics based loss into the neural operator based FWI framework. The experiments on synthetic data showed that the inclusion of the physics loss effectively suppressed noise and led to more stable inversion results. Other applications of neural operators include first arrival picking \citep{sun2023phase}, microseismic event localization \citep{sun2022accelerating}, and traveltime simulation \citep{song2024seismic}. Another exciting category of deep learning algorithms is generative models, particularly diffusion models \citep{ho2022video,peebles2023scalable}, which have attracted increasing attention in recent years. As a new class of generative models, diffusion models differ from traditional approaches such as GANs \citep{goodfellow2020generative} in that their training process is highly stable, and during the generation stage, they can produce high-resolution diverse samples. Owing to these advantages, conditional diffusion models can generate specific samples under given conditions or constraints, thereby serving as a learning paradigm capable of learning the distribution of desired outputs (solutions). For example, in geophysical applications, \cite{cheng2025generative} proposed a supervised conditional diffusion model to generate frequency domain wavefields, where the conditioning information consists of the background wavefield and the corresponding velocity model, and the model subsequently generates the scattered wavefield. Several examples demonstrated that this diffusion model achieves more accurate predictions than conventional supervised learning algorithms. Another distinctive application of the conditional diffusion model lies in its use as a regularization term to constrain the FWI process. \cite{wang2023prior} incorporated the prior provided by a pretrained diffusion model to regularize the FWI process. The field data example demonstrated that the diffusion model effectively injects the prior information learned from the training data into the final inversion results, thereby yielding high resolution velocity building results. \par

{In this paper, we present a novel inversion framework that integrates a neural operator with a diffusion model for high resolution velocity model building. We leverage a neural operator to learn the complex mapping from velocity models to time-lag RTM images, thereby enabling a fast and differentiable approximation to the \textcolor{black}{modeling following by the operator.} In the proposed framework, the neural operator is first pretrained on a large set of synthetic data to accurately predict time-lag RTM images from given velocity models. During the inversion stage, the pretrained neural operator serves as a surrogate model, and automatic differentiation is employed to iteratively update the velocity model by minimizing the misfit between the predicted and observed time-lag RTM images. To further improve the inversion quality, we incorporate a diffusion model as a regularization \textcolor{black}{term}. By conditioning the inversion on the \textcolor{black}{diffusion model learned priors}, the framework effectively suppresses artifacts and enhances the recovery of high frequency components, enabling the reconstruction of high resolution and geologically plausible velocity models. \textcolor{black}{We will show these features on synthetic and real datasets.}

\section{Methods}
In this section, and since we aim to use supervised learning, we first describe the training synthetic velocity models and the corresponding time-lag images. Then, we propose the forward and inverse modeling framework based on a neural operator, which provides a unified and efficient approach for learning the mapping between velocity models and RTM images. Finally, we elaborate on the role of the diffusion model within the overall framework, where it serves as a powerful generative prior to regularize the inversion process and inject useful information into the reconstructed (inverted) velocity models.

\subsection{Training data and the time-shift imaging condition}

Operator learning can also be regarded as a form of supervised learning; therefore, the quality of the training data directly determines the performance of inversion on field data. Given that the subsurface structure is generally horizontally layered (flat), we generate synthetic velocity models that closely resemble realistic geological structures to train the network. For a more detailed description of the dataset generation workflow and preprocessing procedures, readers can refer to \citep{ovcharenko2022multi}. Here, we employ time-lag RTM because it provides richer high wavenumber velocity information from the reflectivity and low wavenumber information from the time lag that is particularly valuable during the inversion stage. The extended image is obtained by cross-correlating the forward- and backward-propagated wavefields with a temporal shift, following the formulation of the time-lag imaging condition \citep{sava2006time}:
\begin{equation}
R(\mathbf{r}, \tau) = \int p^{+}(\mathbf{r}, t + \tau)\, p^{-}(\mathbf{r}, t - \tau)\, dt,
\end{equation}
where \(p^{+}(\mathbf{r}, t)\) and \(p^{-}(\mathbf{r}, t)\) denote the source (forward propagated) and receiver (backward propagated) wavefields, respectively, and \(\tau\) represents the time shift between them. The resulting function \(R(\mathbf{r}, \tau)\) quantifies the degree of temporal correlation between the two wavefields at each spatial location. When \(\tau = 0\), the conventional imaging condition is recovered, producing standard migrated images. In contrast, nonzero time lags (\(\tau \neq 0\)) reveal the residual misalignment of the wavefields caused by velocity inaccuracies, thereby providing valuable information for velocity model updating in the inversion stage. Furthermore, in conventional velocity model building, time-lag RTM (also known as the extended imaging condition) has been widely used to assess the accuracy of the migration velocity model \citep{yang20133d}. Motivated by these properties, we utilize time-lag RTM images as training samples in our neural operator based framework. \par

In Figure~\ref{fig1}(a), we illustrate the entire workflow from the forward modeling of time-lag RTM images to the training of the neural operator. During the training data generation stage (Figure~\ref{fig1}(a)), seismic data are simulated using the true velocity model. In the forward modeling process, 460 receivers recording from 116 sources are both evenly deployed along the surface. For each velocity model, the migration velocity model is obtained by applying a Gaussian smoothing operator with a standard deviation of 15 grid points. Then, the migrated velocity model is used to migrate the simulated seismic data to produce the time-lag RTM images. Figure~\ref{fig2} illustrates representative examples of the synthetic velocity models and their corresponding time-lag RTM images. Next (Figure~\ref{fig1}(a)), we follow a standard supervised training procedure. The neural operator produces its output, which is then compared with the forward modeled time-lag RTM images to compute the loss. The residuals are backpropagated through the network to update the network parameters until convergence. Figure~\ref{fig1}(b) illustrates the workflow of the inversion stage. Note that the channel corresponding to the true velocity model in the training phase is now replaced by the migration velocity, which we assume is available as we also use to obtain the time-lag migrated RTM images from the observed data. During inversion, the process can be conceptually understood as if the seismic data were generated using the migration velocity instead of the true velocity, which in this case corresponds to the field data (unknown). These data are then combined with the second channel (the migration velocity), which was used to perform the migration operation. In other words, we anticipate that the neural operator as a surrogate to the model-migration operations will initially produce images reflecting modeled data from a migration velocity (practically no reflections), and the misfit with the images corresponding to observed data will be used to update the migration velocity to one that reflects the true modeling velocity (true Earth). The output is compared with the observed time-lag RTM images to compute the loss, which is then backpropagated to update the first channel. The inversion process then proceeds by iteratively updating the migration velocity in the first channel so as to progressively reduce the mismatch between the simulated (from the pre-trained neural operator) and observed data. A more detailed explanation of the inversion procedure is presented in subsection 2.

\begin{figure}
\centering
\includegraphics[width=\textwidth]{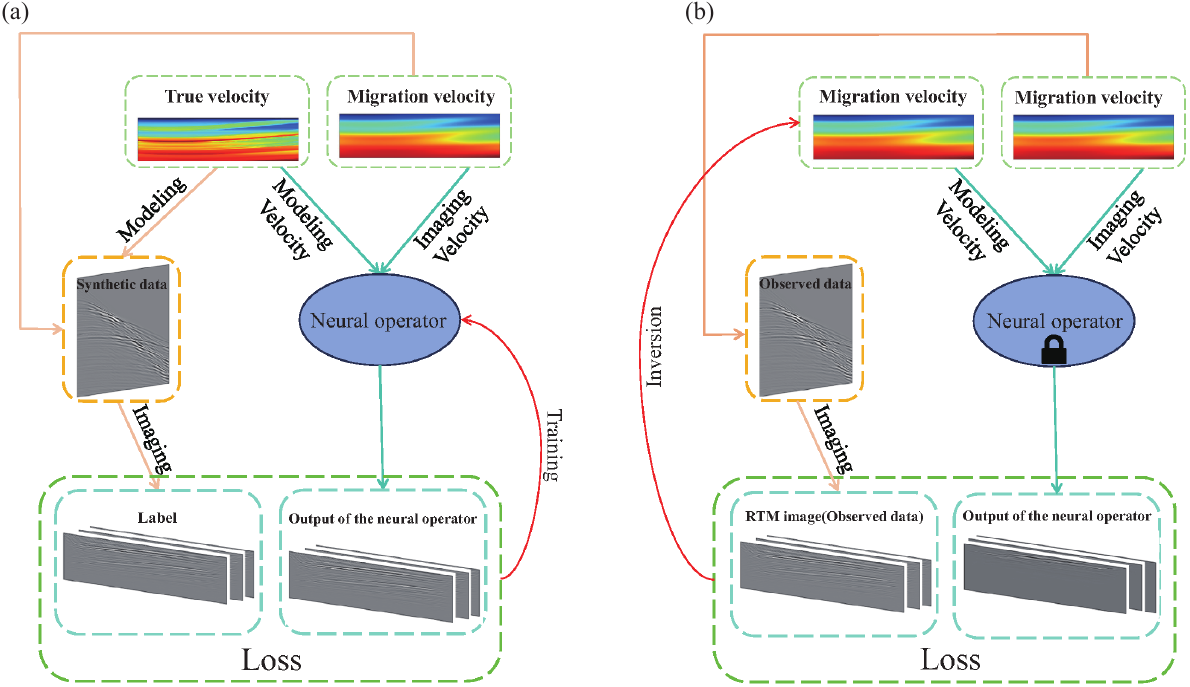}
\caption{ Schematic of the proposed workflow. (a) The data preparation and training stage: Seismic data are simulated using the true velocity model. The migrated velocity model is obtained by applying a Gaussian smoothing and is combined with the simulated data to generate time-lag RTM images. The neural operator is trained in a supervised manner to predict these images, with gradients backpropagated to minimize the loss. (b) The inversion stage: the channel of the true velocity model is replaced by the migration velocity. The pre-trained neural operator is used to predict the time-lag RTM images. The loss between simulated and observed data (images) is backpropagated to update the migration velocity iteratively.}
\label{fig1} 
\end{figure}

\begin{figure}
\centering
\includegraphics[width=\textwidth]{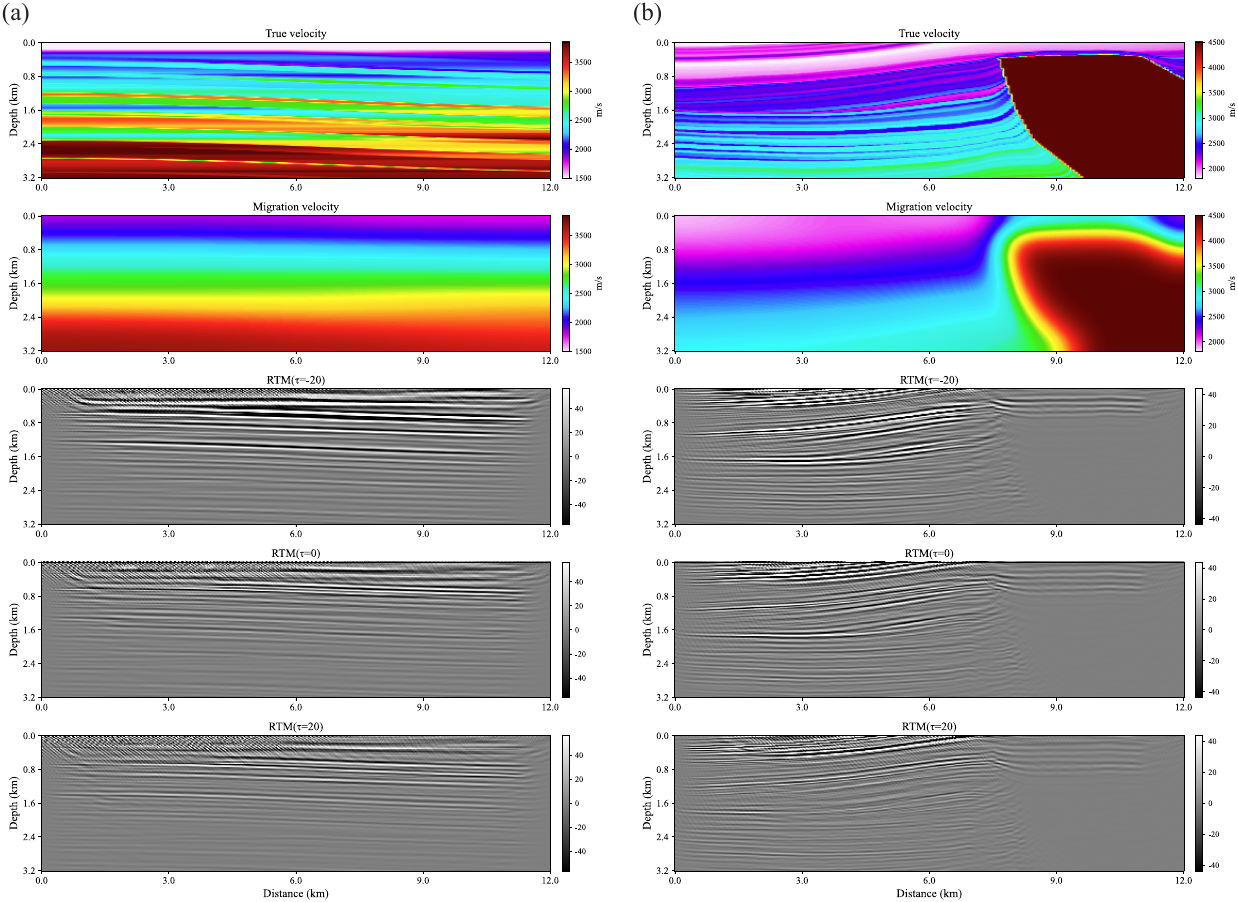}
\caption{ The two columns (a and b) represent two sampled velocity models used during neural network training. From top to bottom, each row corresponds to the true velocity model, the migration velocity model, and the associated time-lag RTM images, respectively.}
\label{fig2} 
\end{figure}

\begin{figure}
\centering
\includegraphics[width=\textwidth]{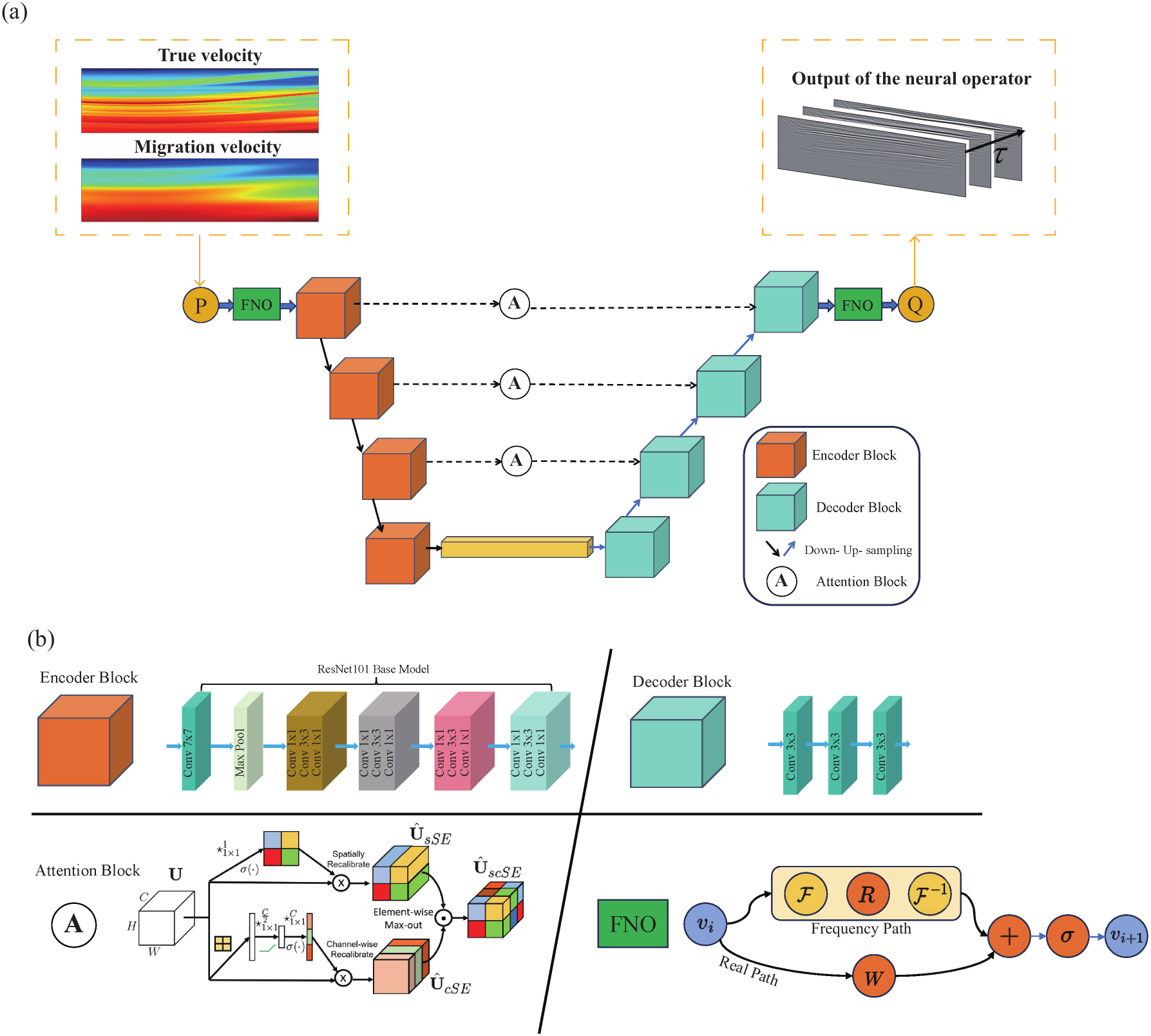}
\caption{The detailed neural operator architecture. The input of the neural operator consists of two channels: the first channel corresponds to the true velocity model, and the second channel represents migration velocity. The output of the operator is the time-lag RTM images. The input data are first processed by an FNO layer, which captures the global spectral representation of the inputs. Subsequently, the features are passed through four encoder blocks and four decoder blocks to extract and reconstruct the hierarchical spatial features. Finally, the resulting feature maps are fed into another FNO layer to produce the final prediction. Each encoder block is built upon the ResNet-101 backbone, which facilitates efficient feature extraction through residual connections. The decoder block consists of three convolutional kernels that progressively reconstruct the spatial details of the feature maps during the upsampling process. During the skip connection stage, we incorporate an attention block that combines both channel attention and spatial attention mechanisms. The detailed architecture of each block is illustrated in (b).}
\label{fig3} 
\end{figure}

\subsection{Operator learning and Automatic differentiation}

Operator learning aims to approximate mappings between infinite-dimensional function spaces rather than finite-dimensional vectors, thereby providing a unified and powerful framework for solving parametric partial differential equations (PDEs). 
Formally, the objective is to learn an operator 
\(\mathcal{G}: \mathcal{A} \rightarrow \mathcal{U}\) 
that maps an input function \(a(\mathbf{x}) \in \mathcal{A}\) (in our case, high resolution, presumbly true, and migration velocity models) to an output function \(u(\mathbf{x}) \in \mathcal{U}\) (in our case, time-lag RTM images), i.e.,
\begin{equation}
    u(\mathbf{x}) = \mathcal{G}(a)(\mathbf{x}), \quad \mathbf{x} \in \Omega,
\end{equation}
where \(\Omega \subset \mathbb{R}^d\) denotes the spatial domain. In practical applications, the operator \(\mathcal{G}\) often represents the solution map of a PDE, for instance:
\begin{equation}
    \mathcal{F}(a, u) = 0,
\end{equation}
where \(\mathcal{F}\) is a differential operator defining the underlying physical law (e.g., the Helmholtz or wave equation). 
The goal of operator learning is thus to approximate \(\mathcal{G}\) directly from data pairs \(\{a_i(\mathbf{x}), u_i(\mathbf{x})\}_{i=1}^N\), such that $\mathcal{G}_\theta \approx \mathcal{G}$, where $\mathcal{G}_\theta$ is parameterized by a neural operator. In particular, we employ the FNO  to approximate the solution operator \(\mathcal{G}\) introduced above. Unlike conventional convolutional neural networks that operate in the spatial domain, FNO learns the integral kernel of the operator directly in the Fourier domain, allowing for efficient representation of global dependencies. \par

In this study, we design a hybrid neural operator architecture that integrates FNO with CNN. Specifically, as shown in Figure~\ref{fig3}, we employ FNO layers at both the encoder (input) and decoder (output) stages of the network to efficiently capture global correlations in the data. The central part of the architecture is constructed using a U-Net structure, where each convolutional block is implemented as a ResNet-101 module. This design allows the model to simultaneously exploit global spectral information through the FNO layers and multi-scale local features through the deep residual U-Net. The neural operator is trained to learn the mapping between velocity models and their corresponding time-lag reverse time migration (RTM) images. The input to the network consists of two channels: the high-wavenumber (considered true, and is sharp) velocity model, which contains fine structural details and is responsible for the observed data, and the low-wavenumber (migration) velocity model, which represents the smooth component of the subsurface structure obtained from conventional velocity analysis and used to obtain the time lag images from the observed data. These two components are concatenated as two-channel inputs to the neural operator. The output of the network is the corresponding time-lag RTM images. By learning this mapping, the neural operator effectively approximates the forward modeling and imaging operations, with velocities responsible for the two operations used as input, enabling efficient generation of extended images that can be further used for inversion and velocity model updating. \par

Figure~\ref{fig4} shows the inversion workflow. During the inversion stage, we keep the pretrained neural operator fixed and employ automatic differentiation to compute the gradient of the extended image with respect to the velocity model. It is worth noting that, unlike the training phase, where the high-wavenumber (sharp) velocity model serves as one of the input channels, in the inversion phase, this channel is initially replaced by the assumed known migration velocity model. By iteratively updating the migration velocity model to minimize the discrepancy between the neural operator’s output and the corresponding target (given by the extended images obtained from migrating the observed data using the migration velocity), we progressively add high wavenumber components from the images to the migration velocity. 
In other words, the inversion is trying to find the velocity model responsible for the forward modeling part of the observed data embedded in the neural operator through training, which eventually corresponds to the true velocity. Formally, the inversion objective can be written as
\begin{equation}
    \mathcal{L}_{\mathrm{inv}}(\mathbf{v}) 
    = \|\, \mathcal{G}_{\theta}^{\mathrm{NO}}(\mathbf{v}) - I_{\mathrm{label}} \,\|_2^2,
\end{equation}
where \(\mathcal{G}_{\theta}^{\mathrm{NO}}\) denotes the pretrained neural operator, \(\mathbf{v}\) is the current velocity model in the first channel, and \(I_{\mathrm{label}}\) represents the target time-lag RTM images. 
The gradient of this loss with respect to the velocity model is then obtained via automatic differentiation:
\begin{equation}
    \nabla_{\mathbf{v}} \mathcal{L}_{\mathrm{inv}} 
    = \frac{\partial \mathcal{G}_{\theta}^{\mathrm{NO}}(\mathbf{v})}{\partial \mathbf{v}}^{\!\top}
      \big( \mathcal{G}_{\theta}^{\mathrm{NO}}(\mathbf{v}) - I_{\mathrm{label}} \big).
\end{equation}
By repeatedly fitting the neural operator’s output to the target extended images and updating the velocity model according to this gradient, the inversion process converges toward a high resolution velocity inversion result.

\begin{figure}
\centering
\includegraphics[width=\textwidth]{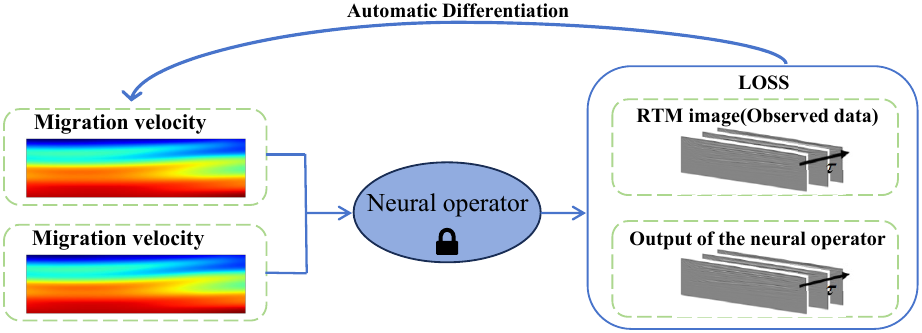}
\caption{In the inversion phase, in the first input channel, we use the migration velocity model, and the neural operator is kept fixed to serve as a differentiable forward model (operator). The network output is compared with the observed seismic data, and gradients are backpropagated via automatic differentiation to iteratively update the velocity model. We continue the iterative process until the predicted images converge to the observed ones.}
\label{fig4} 
\end{figure}

\subsection{Neural-operator-conditioned DDPM}

Motivated by the recent success of diffusion models in geophysical inversion problems \citep{taufik2025conditionally}, we employ a Conditional Denoising Diffusion Probabilistic Model (DDPM) with a U-Net based denoiser, conditioned on the output of the inverse mode neural operator (Figure~\ref{fig4}), to reconstruct high wavenumber velocity components from observed data (RTM images). The key idea of our approach is to use the velocity model inverted by the neural operator as a physical prior to guide the denoising diffusion process. 

\paragraph{(i) Unconditional training.}
In the first stage, we train a standard DDPM in an unconditional manner to learn the underlying distribution of high resolution synthetic velocity models. This unconditional training enables the diffusion model to capture general structural patterns and high wavenumber characteristics of the velocity models without any explicit physical conditioning. 
Formally, the forward diffusion process progressively adds Gaussian noise to a clean velocity model $v_0$:
\begin{equation}
    q(v_{s}\,|\,v_0) = 
    \mathcal{N}\!\left(
        v_{s};
        \sqrt{\bar{\alpha}_{s}}\,v_0, 
        (1 - \bar{\alpha}_{s})\,I
    \right),
\end{equation}
where $s \in \{1, \ldots, S\}$ is the diffusion step, $\bar{\alpha}_{s} = \prod_{i=1}^{s}(1 - \beta_i)$, and $\beta_i$ follows a linear noise schedule. 
The denoiser $\epsilon_{\theta}(v_{s}, s)$ is trained to predict the added Gaussian noise using the simplified DDPM objective:
\begin{equation}
    \mathcal{L}_{\mathrm{simple}}(\theta) 
    = \mathbb{E}_{s,\,v_0,\,\epsilon}
    \Big[
        \big\|
        \epsilon - 
        \epsilon_{\theta}\!\big(
            \sqrt{\bar{\alpha}_{s}}\,v_0 
            + \sqrt{1-\bar{\alpha}_{s}}\,\epsilon, 
            s
        \big)
        \big\|_2^2
    \Big],
\end{equation}
where $\epsilon \sim \mathcal{N}(0, I)$.

\paragraph{(ii) Neural operator guided generation}

During inversion, we estimate the velocity by fitting the operator output to the target RTM label $I_{\mathrm{label}}$:
\begin{equation}
    \mathbf{v}^\star
    \;=\;
    \arg\min_{\mathbf{v}}
    \;\mathcal{L}_{\mathrm{inv}}(\mathbf{v})
    \;=\;
    \arg\min_{\mathbf{v}}\;
    \big\|\,\mathcal{G}_{\theta}^{\mathrm{NO}}(\mathbf{v})-I_{\mathrm{label}}\,\big\|_2^2,
\end{equation}
where gradients $\nabla_{\mathbf{v}}\mathcal{L}_{\mathrm{inv}}$ are obtained via automatic differentiation through the fixed operator $\mathcal{G}_{\theta}^{\mathrm{NO}}$ to update $\mathbf{v}$.

In the subsequent conditional generation phase, conditioning is imposed via initialization. The current inverted velocity model obtained from the inversion stage, $\mathbf{v}$, seeds the reverse process,
\begin{equation}
    v_s
    \;=\;
    \sqrt{\bar{\alpha}_S}\,\mathbf{v}
    \;+\;
    \sqrt{1-\bar{\alpha}_S}\,\epsilon,
    \qquad \epsilon\sim\mathcal{N}(0,I),
\end{equation}
and the denoiser iteratively refines $v_s$:
\begin{equation}
    v_{s-1}
    \;=\;
    \frac{1}{\sqrt{\alpha_s}}
    \left(
        v_s
        -
        \frac{1-\alpha_s}{\sqrt{1-\bar{\alpha}_s}}\,
        \epsilon_{\theta}(v_s,s)
    \right)
    \;+\;
    \sigma_s\,z,
    \quad
    \sigma_s=\sqrt{\beta_s},\;
    z\sim\mathcal{N}(0,I)\ ,
\end{equation}
Subsequently, the generated velocity sample $v_{s-1}$ is fed back into the neural operator based inversion framework for the next iteration, thereby forming a loop process in which the diffusion model and the neural operator iteratively enhance the model to achieve reliable reconstruction of the subsurface velocity. In our inversion framework, since the sampling is performed in a single step, we set the number of diffusion steps to one ($s$=1).
\begin{figure}
\centering
\includegraphics[width=\textwidth]{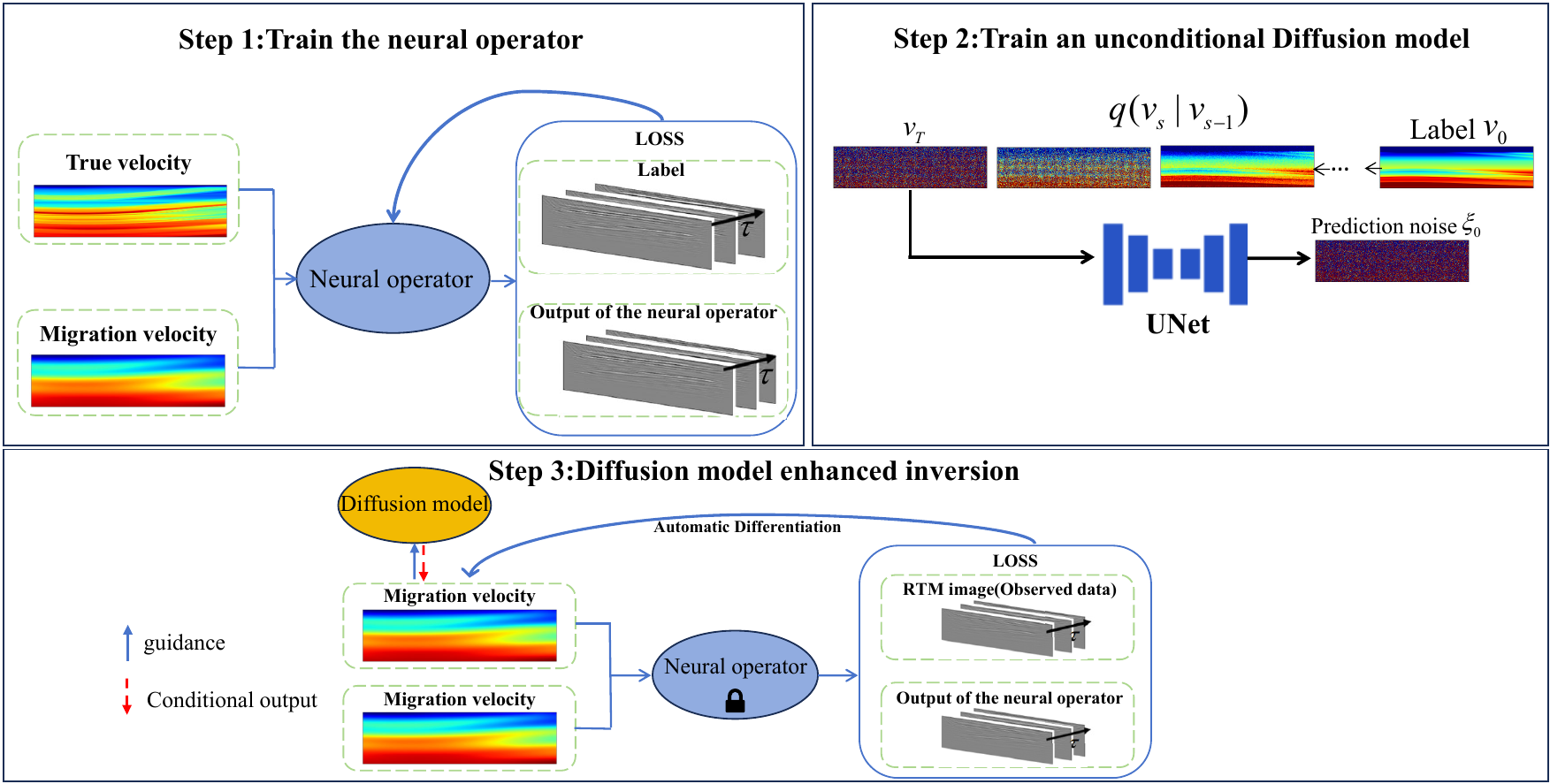}
\caption{Schematic illustration of the Neural-Operator-Guided DDPM framework.
First, the neural operator is trained to rapidly predict the time-lag RTM images, serving as a differentiable forward operator. Next, an unconditional diffusion model is trained, which possesses the capability to generate new velocity samples that follow the desired velocity distribution. In the third stage, the diffusion model is integrated into the inversion framework, where the inversion result at each iteration serves as the guidance (condition) for the diffusion model to generate a new velocity model. The generated velocity model is then used as the initial model for the next round of inversion, progressively refining the reconstruction through iterative updates.}
\label{fig5} 
\end{figure}
%
\section{Results}
In this section, we test the neural operator based inversion on both synthetic and field data. First, when only using the neural operator for inversion, the synthetic examples demonstrate that incorporating the time-lag RTM during the inversion stage effectively enhances the recovery of high-frequency components in the background velocity model. Next, we present the inversion results obtained with the additional constraint of the diffusion model (DDPM+NO), which further improves the reconstruction quality. Finally, we show an encouraging result on challenging field data, highlighting the practical potential of the proposed framework.

\subsection{Training details and forward modeling test}
During the training stage, we use 4,000 samples to train the neural operator, while another 1,000 samples serve as the validation set. The grid size of the training velocity models and RTM images is 128 × 480. The network is optimized using an Adam optimizer with a learning rate of 0.001. Training is performed for 400 epochs on a single NVIDIA A100 GPU, taking approximately 6 hours and 34 minutes to complete. Figure~\ref{fig6} shows the training and validation losses. For the diffusion model, we also use an Adam optimizer with a learning rate of $1 \times 10^{-6}$, and we train the diffusion model for 500 epochs to store the distribution of the high resolution velocity models used in training the neural operator, which takes approximately 23 hours.

\begin{figure}
\centering
\includegraphics[width=1\textwidth]{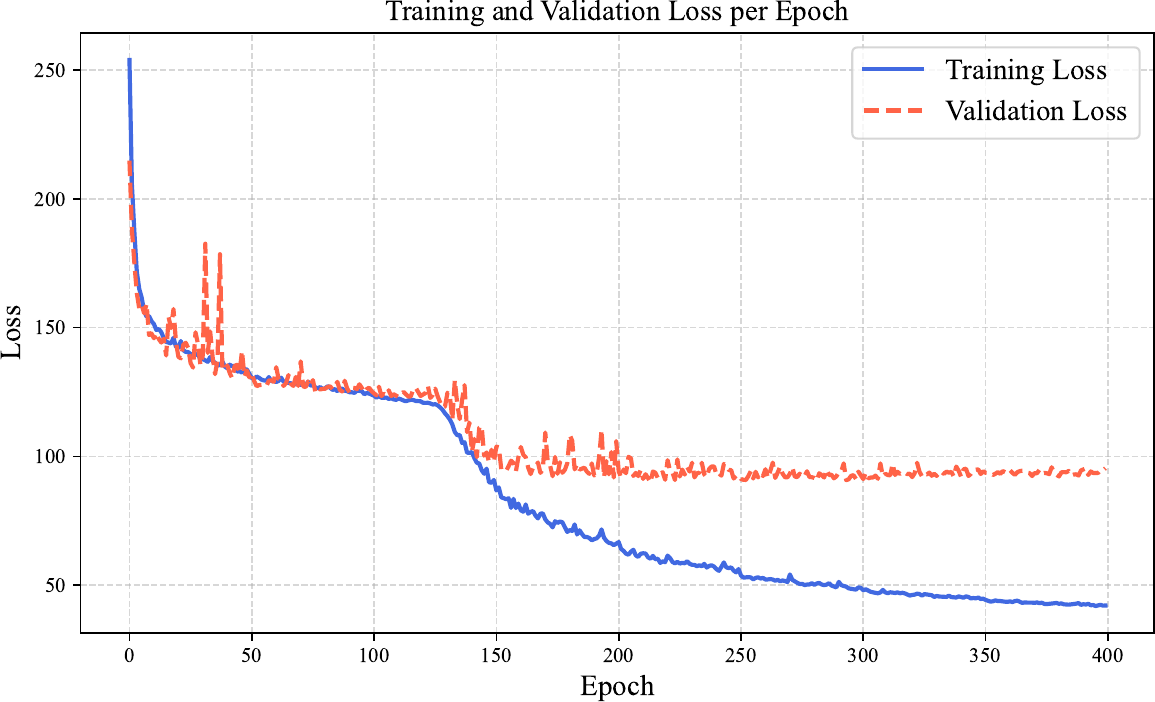}
\caption{Training and validation losses during the training phase of the neural operator.}
\label{fig6} 
\end{figure}
We first evaluate the forward process of the neural operator. To do so, we test the fully trained neural operator using synthetic velocity models from the test set. The forward modeling results are shown in Figure~\ref{fig7}. As illustrated in Figure~\ref{fig7}(g)–(i), the predictions generated by the neural operator are in close agreement with the reference labels (Figure~\ref{fig7}(d)–(f)), demonstrating its high accuracy in reproducing the time-lag RTM images. Moreover, the neural operator produces time-lag RTM output within approximately 0.9 seconds, in contrast to the 4 minutes and 50 seconds required by RTM, achieving an acceleration of nearly 300 times.
\par

\begin{figure}
\centering
\includegraphics[width=1\textwidth]{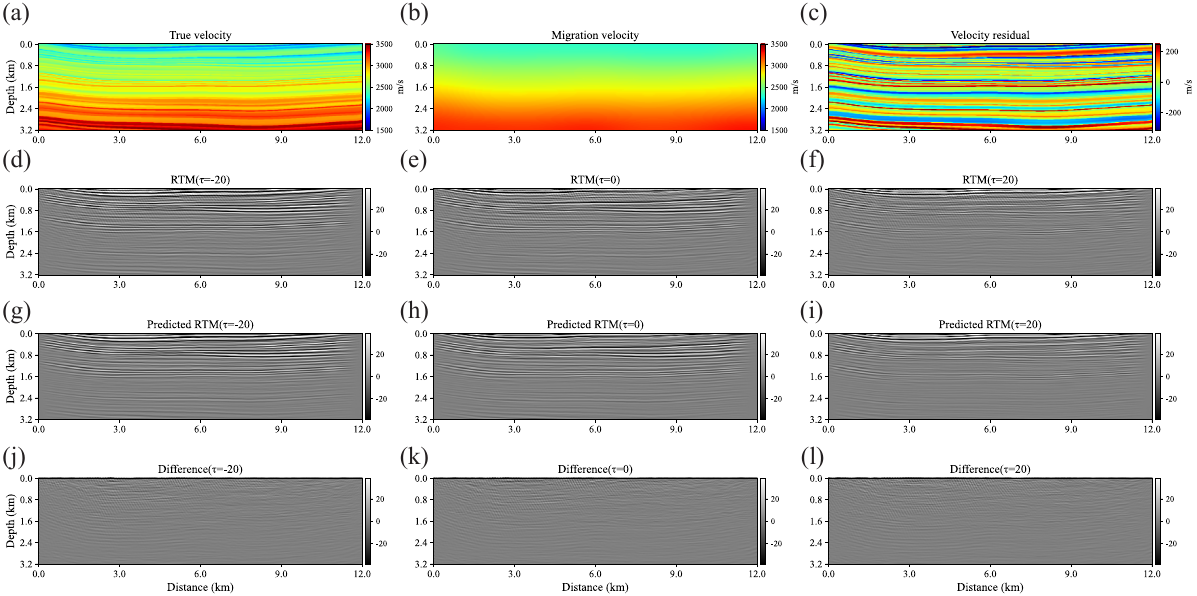}
\caption{Forward propagation test of the neural operator using a sample from the test dataset. (a) and (b) show the neural operator inputs, and their difference (c), (d)–(f) correspond to the ground truth, (g)–(i) present the predicted results, and (j)–(l) display the prediction errors, plotted as the same scale. }
\label{fig7} 
\end{figure}

\subsection{Synthetic data inversion test}

First, without the diffusion model regularization, we test the impact of incorporating the time-lag RTM compared to only using only the zero lag image \citep{chattopadhyay2008imaging}. Specifically, the zero time-lag configuration uses only the $\tau$ = 0 RTM image during both the training and inversion stages. During the inversion stage, we also employ an Adam optimizer to update the velocity model over a total of 300 iterations. The learning rate is gradually decayed by a factor of 0.8 every 100 iterations. The entire inversion process requires approximately 28 seconds to complete. \par
Figure~\ref{fig8} presents the inversion results obtained using only the neural operator-based framework without the diffusion regularization. A comparison between Figures~\ref{fig8}(c) and (d) clearly demonstrates that incorporating the time-lag RTM as observed data significantly enhances the recovery of high-frequency components in the velocity model (near the black arrow in Figure~\ref{fig8}(c)). To further demonstrate the effectiveness of the time-lag RTM images, Figure~\ref{fig9} presents a vertical profile and its corresponding spectrum. In particular, the amplitude differences in the wavenumber domain shown in Figure~\ref{fig9}(b) clearly indicate that the inversion result incorporating time-lag information achieves a better recovery of the mid-wavenumber components. However, it should also be noted that the inversion results obtained using only the neural operator contain considerable artifacts. This issue is likely attributed to the noisy gradients during the backpropagation stage, which can lead to unstable updates in the velocity model. In addition, due to the inherent geometric limitations of the RTM image, the inversion accuracy near the model boundaries is relatively poor. To mitigate these issues and further improve the overall inversion accuracy, a regularization term is therefore required to suppress the unwanted noise components and enhance the inversion result.

\begin{figure}
\centering
\includegraphics[width=1\textwidth]{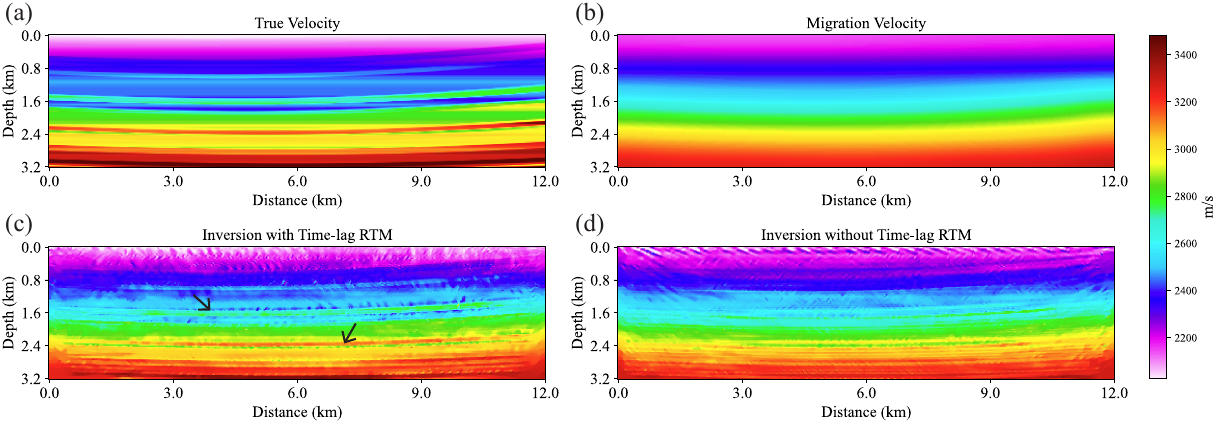}
\caption{Comparison of inversion results using zero-lag image and time-lag RTM images. (a) shows the reference (true) velocity model, (b) illustrates the migration velocity model used in the inversion, (c) presents the inversion result obtained with the time-lag RTM images, and (d) shows the result derived using the zero-lag RTM image.}
\label{fig8} 
\end{figure}

\begin{figure}
\centering
\includegraphics[width=1\textwidth]{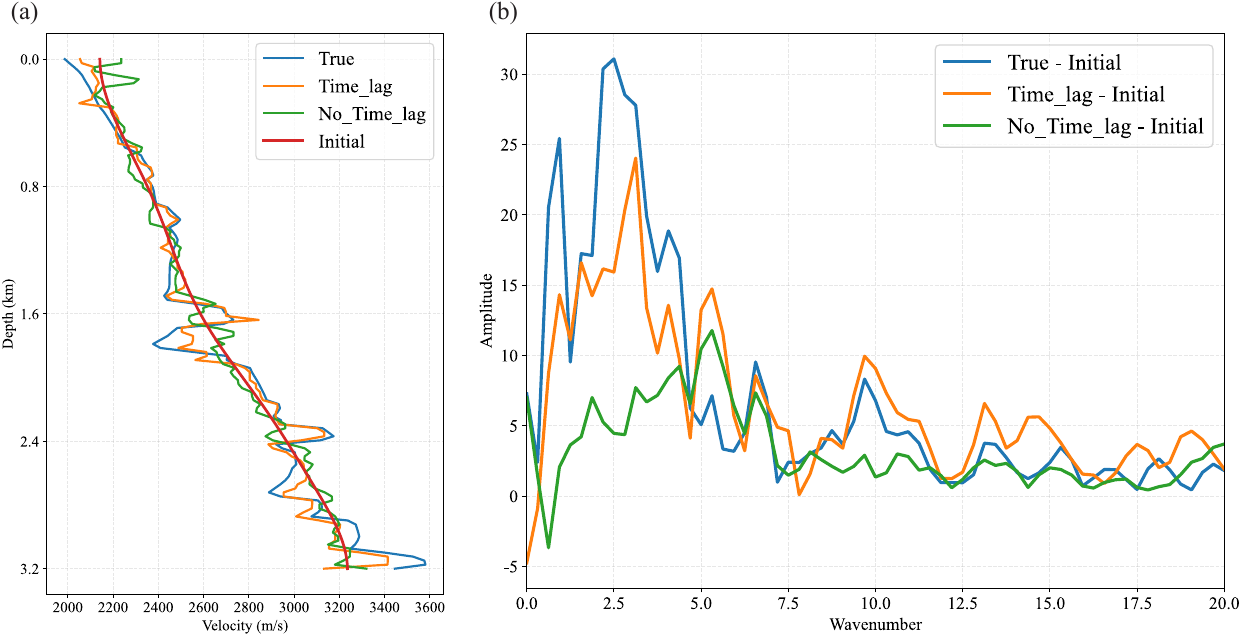}
\caption{(a) The vertical profile from Figure~\ref{fig8} (distance=6.0 km), and (b) shows the amplitude differences in the wavenumber domain relative to the initial velocity.}
\label{fig9} 
\end{figure}

\subsection{Diffusion prior enhanced inversion test}
Before performing the inversion, it is essential to ensure that the unconditional diffusion model is capable of generating high resolution velocity samples. This step guarantees that the pretrained diffusion prior has successfully learned the statistical distribution of high-wavenumber structures, which is crucial for guiding the subsequent conditional inversion processes. 

\begin{figure}
\centering
\includegraphics[width=1\textwidth]{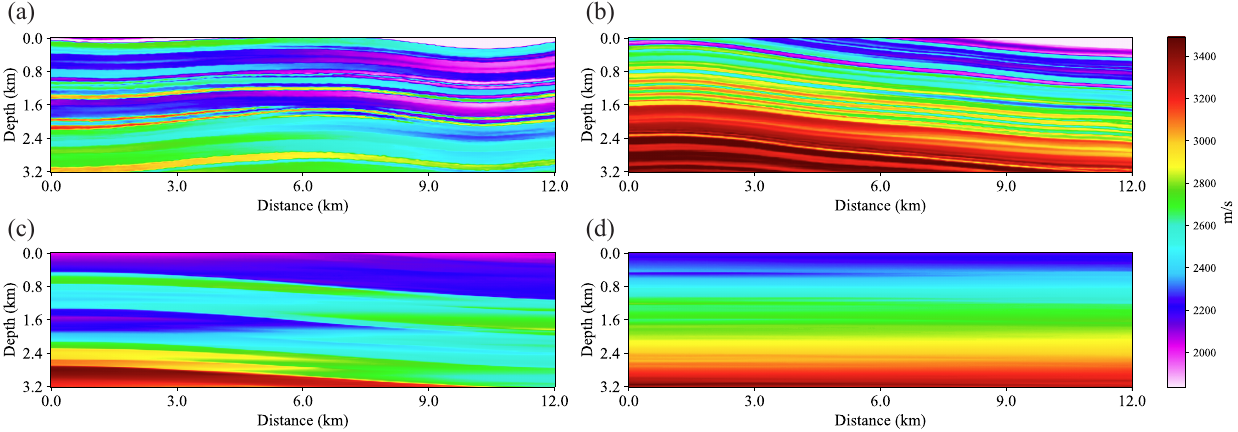}
\caption{Velocity samples unconditionally generated by the diffusion model.}
\label{fig10} 
\end{figure}

Figure~\ref{fig10} displays four randomly generated samples produced by the trained diffusion model. 
We can clearly observe that the diffusion model is capable of synthesizing high-resolution velocity in an unsupervised manner. Next, we proceed with the procedure illustrated in Step~3 of Figure~\ref{fig5}. At each iteration, the velocity model obtained from the neural operator based inversion serves as the guidance (or condition) for the diffusion model to generate new velocity samples. The generated sample, which incorporates both the physics-informed structure from the neural operator and the high wavenumber details modeled by the diffusion process, is then fed back as the current model for the subsequent inversion cycle. It should be noted that, during the reverse process of the diffusion model, we employ only a single step generation, i.e., the diffusion model performs generation in just one time step. This strategy significantly reduces the computational cost. On the synthetic dataset, the inversion without the diffusion model takes approximately 28 seconds, whereas incorporating the diffusion model increases the runtime to about 113 seconds. Figure~\ref{fig11} presents the same example as shown in Figure~\ref{fig8} for a direct comparison. By examining Figure~\ref{fig11}(b) and Figure~\ref{fig11}(c), we can observe that incorporating the diffusion model as a regularization term effectively suppresses inversion artifacts and stabilizes the reconstructed velocity field. Furthermore, within the region highlighted by the black line elipse, the diffusion prior appears to be successfully embedded into the inversion results.

\begin{figure}
\centering
\includegraphics[width=1\textwidth]{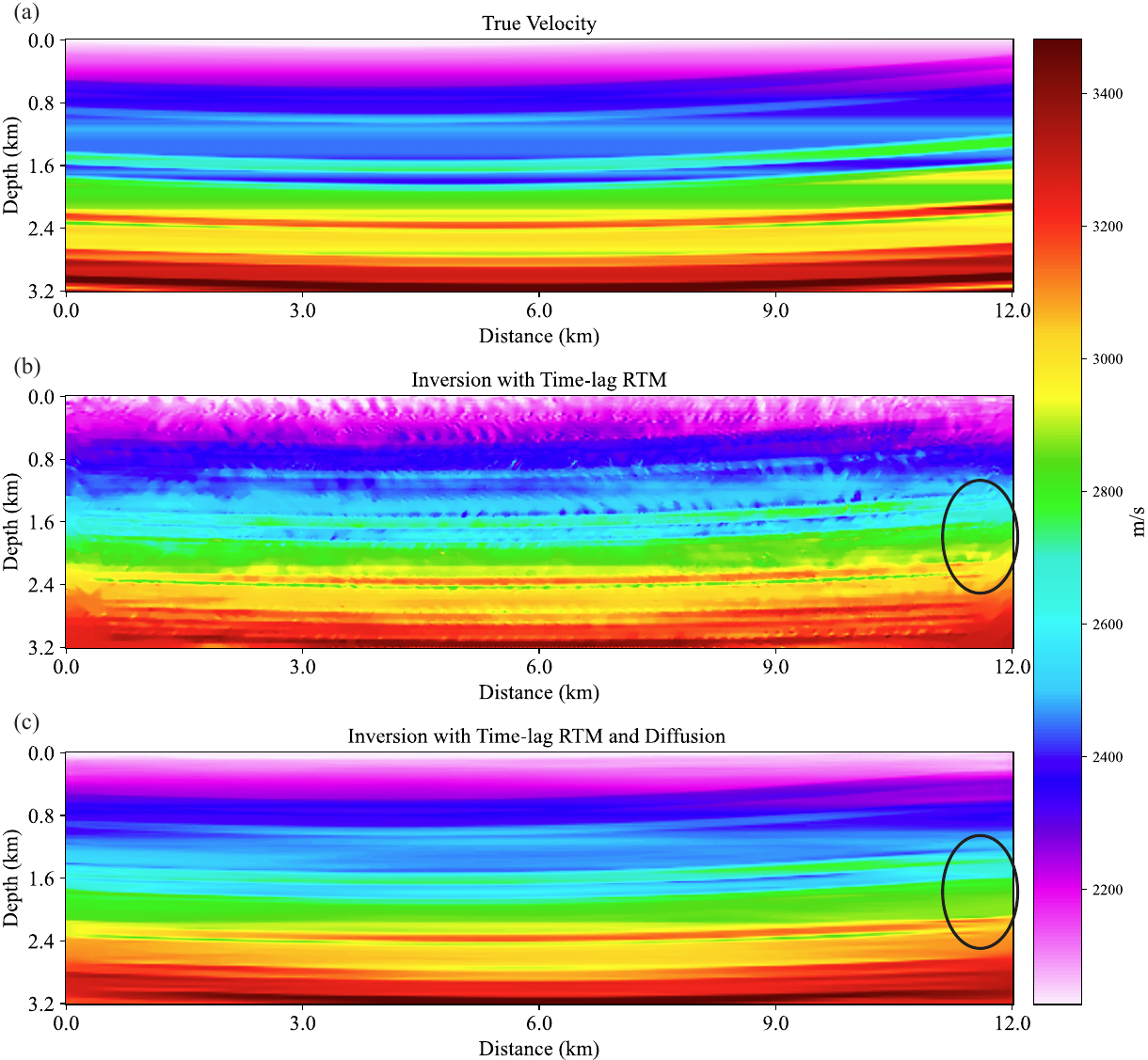}
\caption{Comparison of inversion results with and without the diffusion model. (a) shows the true velocity model (corresponding to Figure~\ref{fig8}(a)), (b) presents the inversion result without the diffusion model (corresponding to Figure~\ref{fig8}(c)), and (c) displays the inversion result obtained with the diffusion model.}
\label{fig11} 
\end{figure}
Figure~\ref{fig12} presents another challenging example. In this case, a prominent salt dome is located in the right half of the model, and the overall velocity structure exhibits sharp contrasts, posing significant difficulties for the inversion process. It is clear that the migration velocity model shown in  Figure~\ref{fig12}(b) lacks high wavenumber components. Figure~\ref{fig12}(c) illustrates the inversion result obtained using only the neural operator, where strong noise artifacts are evident, and deeper structures are poorly recovered. In contrast, when the diffusion model is incorporated, prior information is effectively injected into the inversion process, which not only suppresses spurious artifacts but also enhances the recovery of high resolution geological details (Figure~\ref{fig12}(d)). Furthermore, we compare the observed and predicted data to evaluate the accuracy of the inversion. As shown in Figure~\ref{fig13}, the predicted data closely match the observed time-lag RTM images, indicating that the reconstructed velocity model can reliably reproduce the seismic data. One clear weakness, that would have affected FWI as well, is the inability to properly capture the salt flank. This is a direct result of our inability to image the reflectivity of the salt flank because there was not enough velocity increase with depth in this model to allow us to do so. FWI gradient will also miss such reflectivity.

\begin{figure}
\centering
\includegraphics[width=1\textwidth]{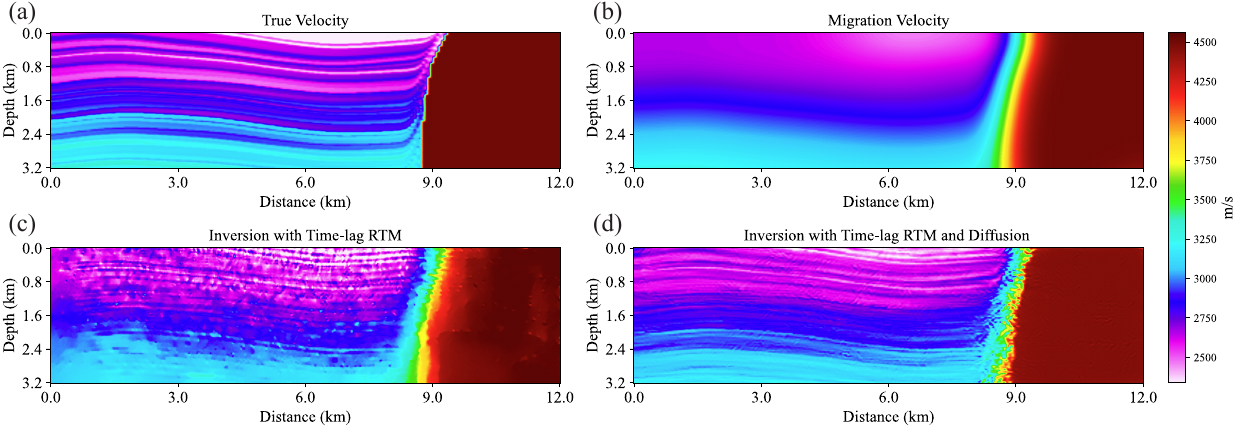}
\caption{Inversion test on a more complex velocity model. (a) and (b) show the true and migration velocity models, respectively. (c) and (d) present the inversion results without and with the diffusion model, respectively.}
\label{fig12} 
\end{figure}

\begin{figure}
\centering
\includegraphics[width=1\textwidth]{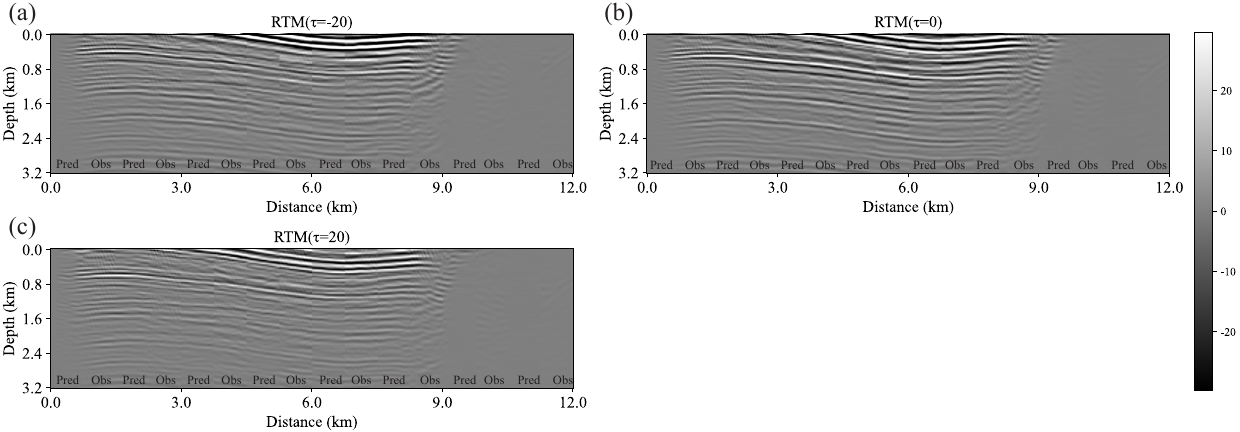}
\caption{Interleaved comparison between the predicted and observed seismic data (Time-lag RTM images).}
\label{fig13} 
\end{figure}

\subsection{Field data test}
Next, we evaluate the proposed method on a field dataset. As shown in the Figure~\ref{fig14}, this dataset represents a challenging real case scenario. The dataset comes from a towed-streamer acquisition acquired by CGG in the North Western part of Australia, consisting of a total of $1,824$ shots and $648$ receivers per shot. The maximum offset is $8.2$ km, and the maximum recording time is $7.0$ s. It spans approximately 30 km in the horizontal direction and about 4 km in depth. We use the migration velocity shown in Figure~\ref{fig14}(a) to obtain the time lag images shown in Figure~\ref{fig14}(b)-(d). These time-lag images are normalized to match the range of the training data. To enable the diffusion model to operate efficiently under such a large-scale setting, we adapt a patch-based inversion strategy, in which the observed seismic data are divided into multiple overlapping patches for inversion. We use the same pretrained neural operator and diffusion model demonstrated in the previous examples. In other words, our neural operator and diffusion model were trained only once for all examples in the paper. The inversion using our proposed framework takes approximately five minutes to complete. By comparing Figure~\ref{fig15}(a) and Figure~\ref{fig15}(b), we find that incorporating the diffusion model effectively suppresses artifacts, and prior information contained in the training data is successfully embedded into the inversion results. We further evaluated the data fitting performance of the zero-lag RTM image. Furthermore, Figure~\ref{fig16} shows that the velocity model obtained from the neural operator recovers a portion of the high frequency information (green curve). Also, when the diffusion model is incorporated, some of the high frequency noise is effectively suppressed (yellow curve). In Figure~\ref{fig17}, the agreement between the predicted and observed data is highly accurate, indicating that the high frequency components embedded in the RTM image are effectively recovered. In addition, we compare the RTM images obtained using the migration (Figure~\ref{fig18}(a)) and inverted (Figure~\ref{fig18}(b)) velocity models. As indicated by the red arrows in Figure~\ref{fig18}, the RTM image produced by our inversion framework exhibits more continuous reflectors, demonstrating that the inverted velocity model provides a better kinematic consistency and improved subsurface imaging quality.

\begin{figure}
\centering
\includegraphics[width=1\textwidth]{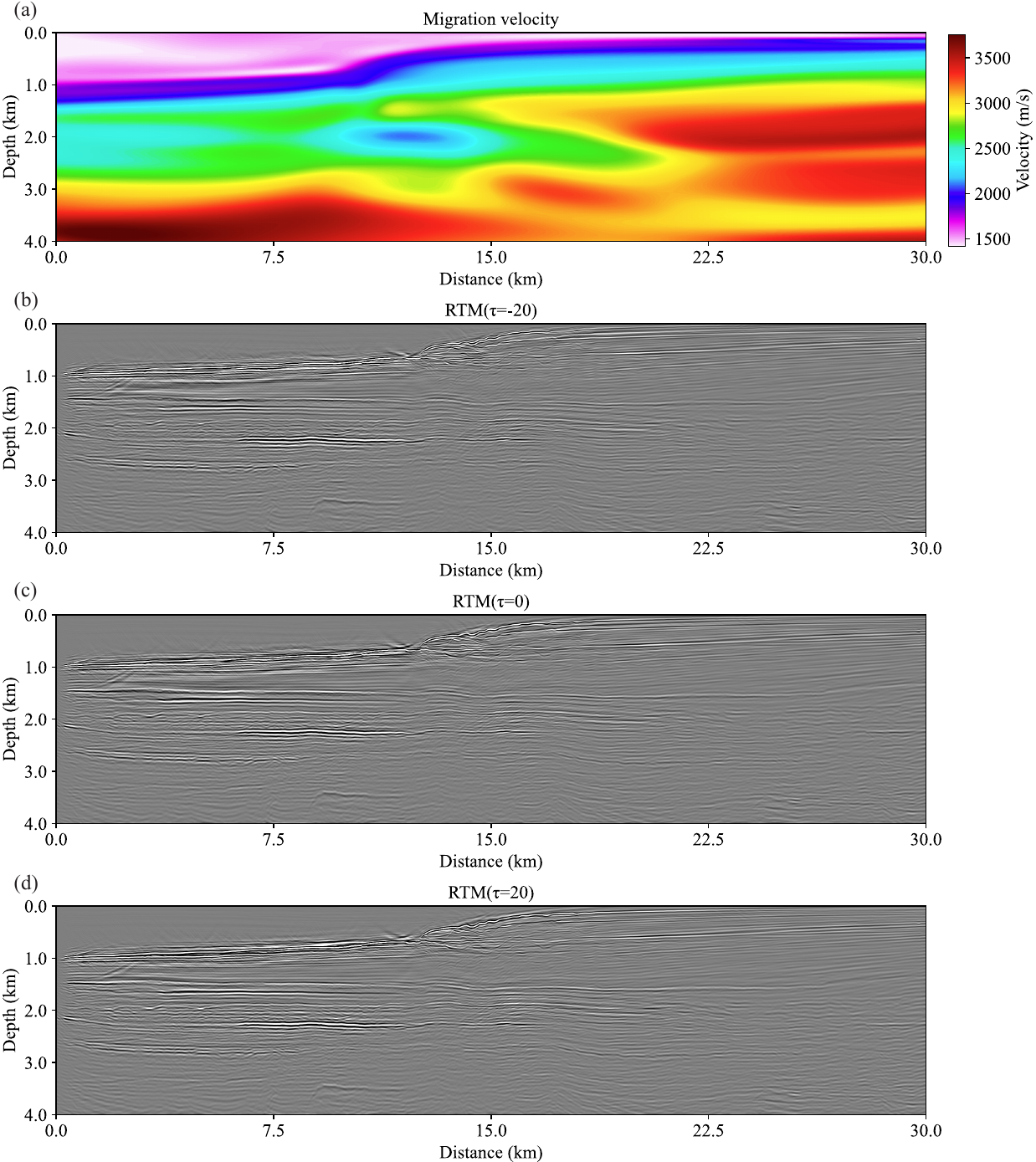}
\caption{(a) shows the migration velocity model, (b)-(d) presents the corresponding time-lag RTM images}
\label{fig14} 
\end{figure}

\begin{figure}
\centering
\includegraphics[width=1\textwidth]{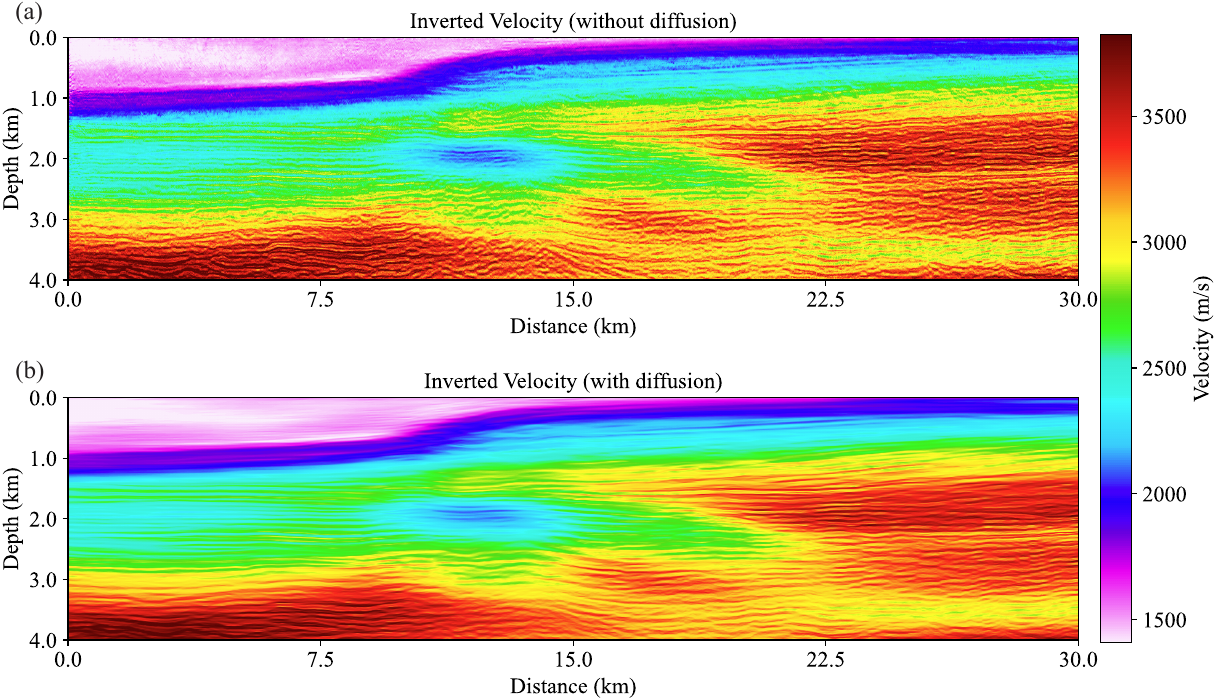}
\caption{(a) and (b) present the inversion results without and with the diffusion model, respectively}
\label{fig15} 
\end{figure}

\begin{figure}
\centering
\includegraphics[width=0.7\textwidth]{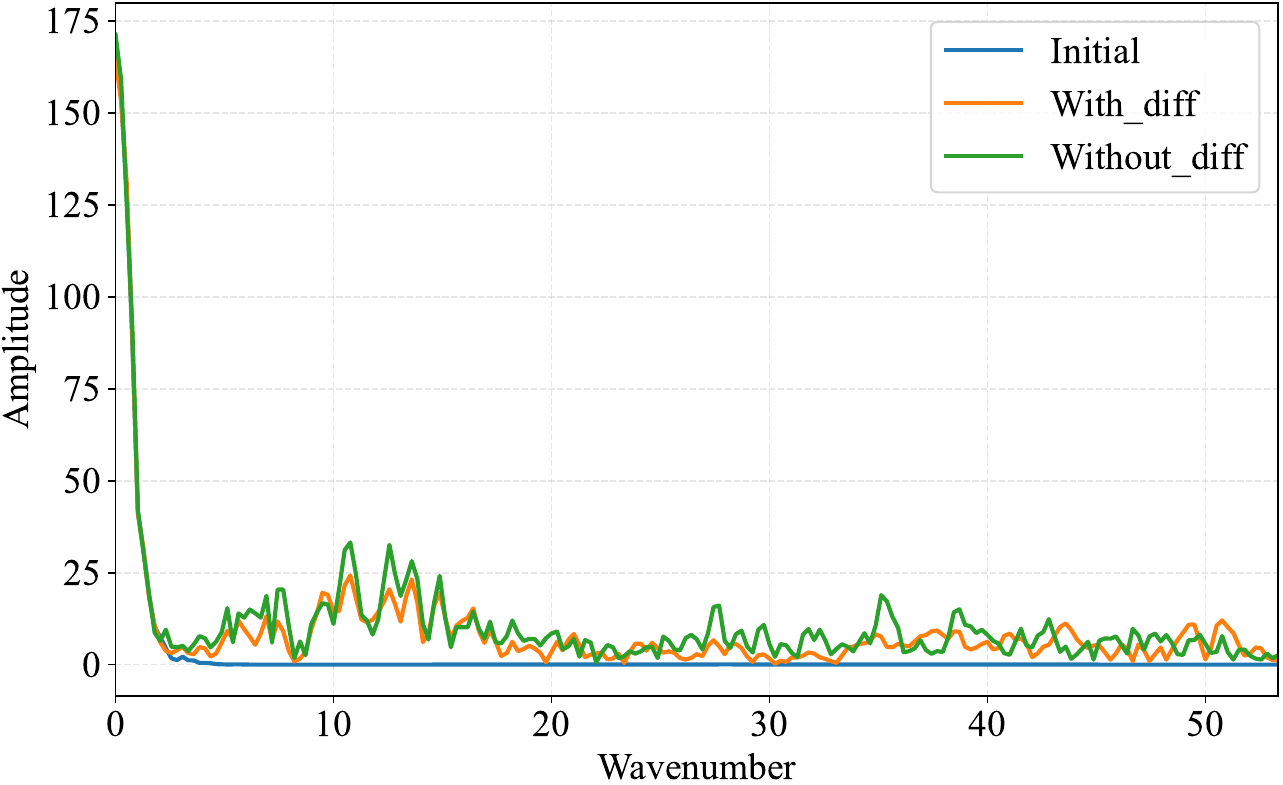}
\caption{The vertical velocity profile spectrum at the location of 16.875 km is shown, where the blue curve corresponds to the initial velocity model (Figure~\ref{fig14}(a)), the yellow curve represents the inversion result obtained with the diffusion model (Figure~\ref{fig15}(a)), and the green curve denotes the inversion result obtained without the diffusion model (Figure~\ref{fig15}(b)).}
\label{fig16} 
\end{figure}

\begin{figure}
\centering
\includegraphics[width=1\textwidth]{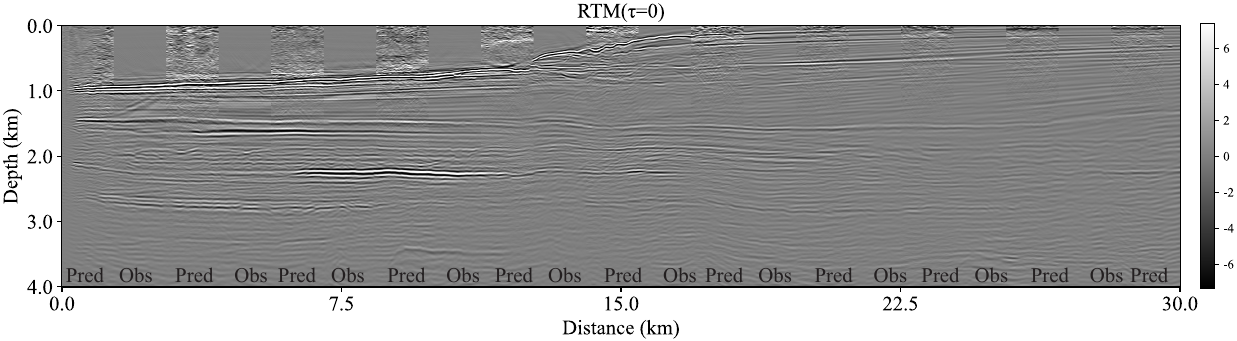}
\caption{Interleaved comparison between the predicted and observed seismic data (Zero-lag RTM image).}
\label{fig17} 
\end{figure}

\begin{figure}
\centering
\includegraphics[width=1\textwidth]{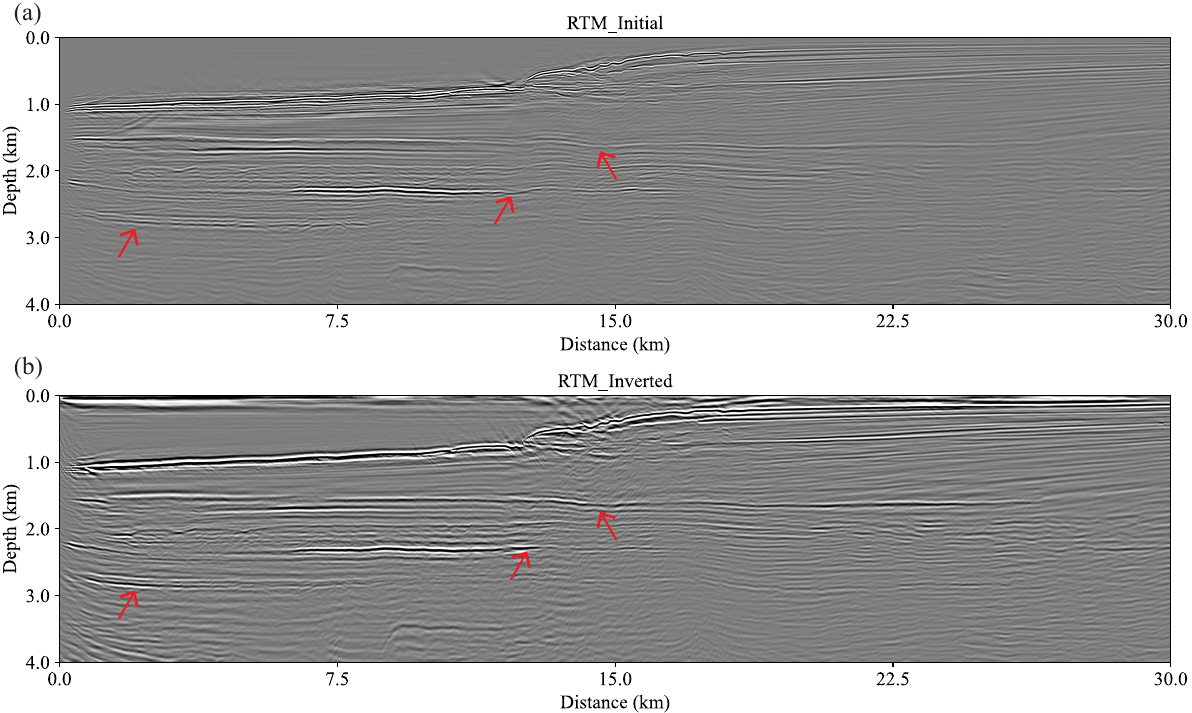}
\caption{RTM images using (a) the migration velocity model (Figure~\ref{fig13}(a)) and (b) the inverted model (Figure~\ref{fig14}(b))}
\label{fig18} 
\end{figure}
\section{Discussion}
In this section, we discuss several key aspects of the proposed inversion framework, as well as potential directions for future improvements.

\subsection{Dependence on the migration velocity model}
As the observed data used in this inversion framework are the time-lag RTM images, the accuracy of the migration velocity model plays a crucial role in the accuracy of the inverted models. Though, as we demonstrated in Figure~\ref{fig8} and Figure~\ref{fig9}, the time lag implementation might add some kinematic (low wavenumber) information, such information is limited. Inaccurate migration velocity models may lead to a wrong inversion result. An inaccurate or overly smooth migration model can distort the kinematic consistency between the predicted and observed RTM images, thereby leading to incorrect updates during inversion and ultimately yielding erroneous results. In our future work, we plan to explore strategies to reduce the dependency on the migration model, such as incorporating data-driven preconditioning, multi-scale inversion schemes. These approaches are expected to enhance the robustness of the inversion framework, enabling it to recover high resolution velocity models even when the migration velocity information is highly uncertain or incomplete. However, for now, we assume that the migration velocity is the velocity that we are comfortable with to do an accurate migration. Usually, such velocities are also used to start FWI.

\subsection{Seismic data matching}

Another way to assess the reliability of the inversion results is to compare the shot gathers obtained from the predicted model with the observed data, like in FWI. This is a big challenge, as fitting the data was never our objective. We will perform this comparison for the field data. As shown in Figure~\ref{fig19}(a), “Init” denotes the data simulated using the migration velocity model (Figure~\ref{fig15}(a)), “Pred” represents the data generated using the inverted velocity model (Figure~\ref{fig15}(b)), and “Obs” corresponds to the field acquired (observed) data. A comparison between Figure~\ref{fig19}(a) and Figure~\ref{fig19}(b) reveals that the predicted data exhibit a better agreement with the observed data, particularly in the diving-wave region, indicating that the low-wavenumber components of the velocity model have been recovered. This improvement is especially evident at far offsets (highlighted by the red arrows). It is important to note that our inversion framework does not include any data matching term in the loss function, implying that enhanced data consistency is an emergent property rather than an explicit optimization objective. More importantly, this interesting, imperfect result is obtained considering that the training samples we used had plenty of limitations in representing the Earth model. We attribute this generally good result in data fitting to the inversion implementation of this velocity model building procedure, which also allowed us to incorporate a diffusion regularization.

\subsection{The training set}

The training set used for machine learning applications is crucial to its success. In this paper, we utilized a single training set (velocity models) for our neural operator training and Diffusion model training for all the examples shown. The training set was reasonably constructed and included many features. However, more experienced practitioners would be able to assemble far more realistic and feature intensive training sets. The training set, thanks to the Diffusion regularization, also serves as a prior to guide the solution in fitting our Earth exceptions. So we expect much better results when better training samples are used. In addition, our acquisition scenario for the training samples for the neural operator was different than the field data acquisition, and yet the inversion admitted good results, demonstrating a certain level of acquisition-related robustness in our framework. Nevertheless, closing the gap between training and field acquisition remains an important direction for future improvement.

\begin{figure}
\centering
\includegraphics[width=\textwidth]{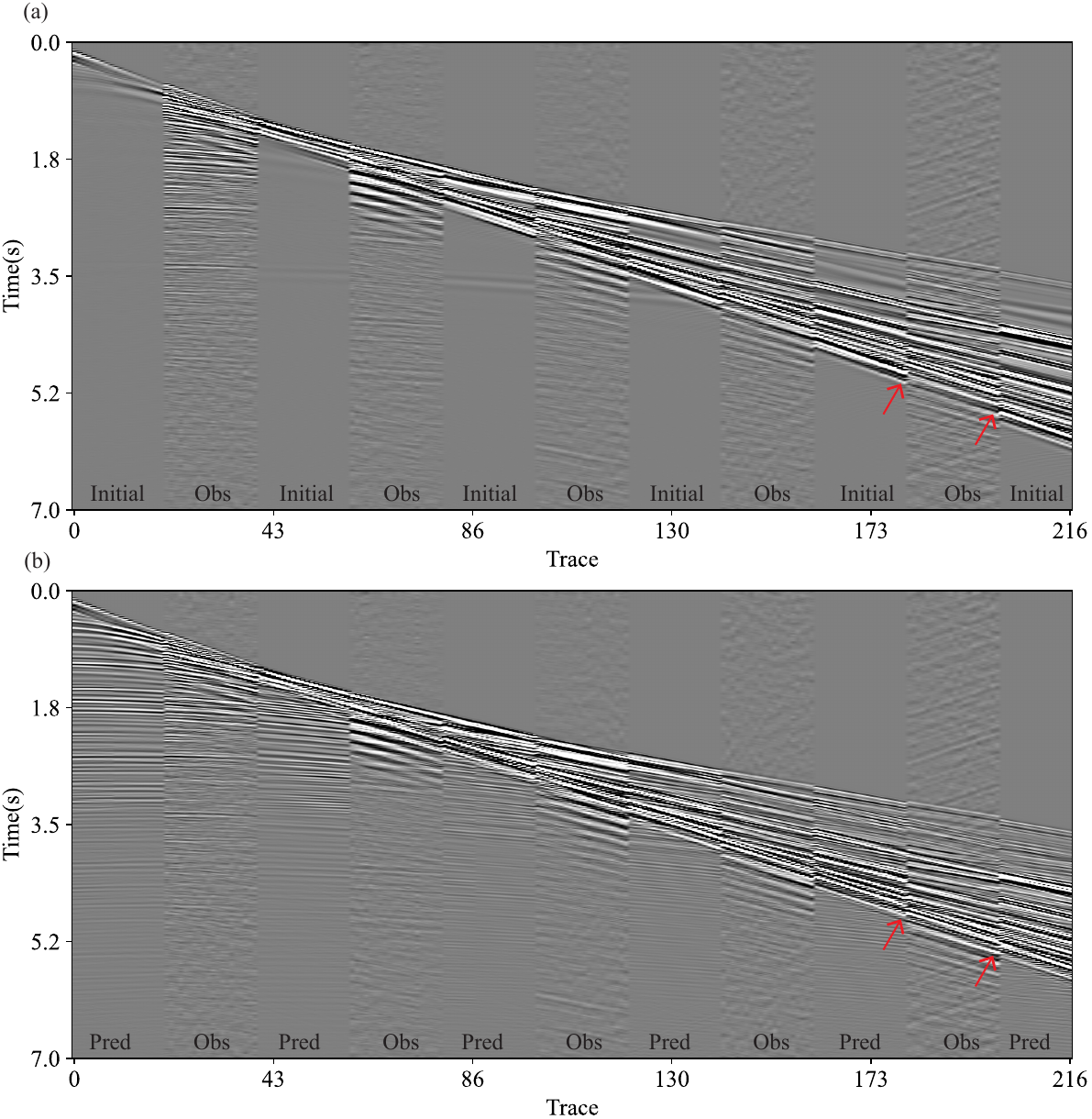}
\caption{Interleaved comparison between the migration, predicted, and observed shot gather. }
\label{fig19} 
\end{figure}
\section{Conclusions}

We proposed a novel inversion framework that integrates a neural operator with a diffusion model regularization. Within this framework, we first train the neural operator using synthetic datasets. The trained neural operator can rapidly generate time lag RTM images, which are then incorporated into an automatic differentiation process to inject the high wavenumber information contained in the time-lag RTM images into the migration velocity model, thereby reconstructing a high resolution velocity model. The experiments demonstrate that introducing the time-lag component significantly enhances the recovery of high frequency features in the velocity model. Subsequently, the diffusion model is incorporated as a regularization term within the inversion process. This regularization effectively suppresses inversion artifacts and introduces prior high frequency information learned from the training data. The results on both synthetic and field datasets confirm that the proposed inversion framework achieves high resolution velocity models more efficiently than conventional FWI methods.\\

\section{\textbf{Acknowledgment}}
This publication is based on work supported by the King Abdullah University of Science and Technology (KAUST). The authors thank the DeepWave sponsors for their support. This work utilized the resources of the Supercomputing Laboratory at King Abdullah University of Science and Technology (KAUST) in Thuwal, Saudi Arabia.
\vspace{0.5cm}
\section{\textbf{Code Availability}}
The data and accompanying codes that support the findings of this study are available at: 
\url{https://github.com/DeepWave-KAUST/VMB_with_diffusion-FNO}. (During the review process, the repository is private. Once the manuscript is accepted, we will make it public.)

\bibliographystyle{unsrtnat}
\bibliography{references}

\end{document}